\documentclass{article}

\usepackage{arxiv}

\usepackage[utf8]{inputenc} % allow utf-8 input
\usepackage[T1]{fontenc}    % use 8-bit T1 fonts
\usepackage{hyperref}       % hyperlinks
\usepackage{url}            % simple URL typesetting
\usepackage{booktabs}       % professional-quality tables
\usepackage{amsfonts}       % blackboard math symbols
\usepackage{nicefrac}       % compact symbols for 1/2, etc.
\usepackage{microtype}      % microtypography
\usepackage{lipsum}
\usepackage{svg}
\usepackage{mdframed}
\usepackage{algorithm}
\usepackage{algpseudocode}
\graphicspath{ {./images/} }

\title{ArxEval: Evaluating the Retrieval and Generation of Language Models for Scientific Papers}

\author{
Aarush Sinha \\
School of Computer Science and Engineering\\
Vellore Institute of Technology-Chennai\\
Chennai - 600127, India \\
\texttt{aarush.sinha2021@vitstudent.ac.in} \\
\And
Dipshikha Chakraborty \\
School of Computer Science and Engineering\\
Vellore Institute of Technology-Chennai\\
Chennai - 600127, India \\
\texttt{dipshikha.chakraborty2021@vitstudent.ac.in} \\
\And
Viraj Virk \\
School of Computer Science and Engineering\\
Vellore Institute of Technology-Chennai\\
Chennai - 600127, India \\
\texttt{viraj.2021@vitstudent.ac.in} \\
\And
Sreeja P. S$^{\dagger}$ \\
School of Computer Science and Engineering\\
Vellore Institute of Technology-Chennai\\
Chennai - 600127, India \\
\texttt{sreeja.ps@vit.ac.in} \\
$^{\dagger}$ Corresponding Author
}

\begin{document}

\maketitle

\begin{abstract}
Language Models [LMs] are now playing an increasingly large role in information generation and synthesis; the representation of scientific knowledge in these systems needs to be highly accurate. A prime challenge is hallucination; that is, generating apparently plausible but actually false information, including invented citations and nonexistent research papers. This kind of inaccuracy is dangerous in all the domains that require high levels of factual correctness, such as academia and education. This work presents a pipeline for evaluating the frequency with which language models hallucinate in generating responses in the scientific literature. We propose \textbf{ArxEval}, an evaluation pipeline with two tasks using ArXiv as a repository: \textbf{Jumbled Titles} and \textbf{Mixed Titles}. Our evaluation includes fifteen widely used language models and provides comparative insights into their reliability in handling scientific literature.

\end{abstract}

% Uncomment the following to link to your code, datasets, an extended version or similar.
%
% \begin{links}
%     \link{Code}{https://aaai.org/example/code}
%     \link{Datasets}{https://aaai.org/example/datasets}
%     \link{Extended version}{https://aaai.org/example/extended-version}
% \end{links}

\section{Introduction}

Large Language Models (LLMs) have emerged as pivotal tools in information access and generation, particularly through their capabilities of producing factually accurate texts. As these models become increasingly integrated into various applications, ensuring the accuracy of their responses has become very important. The performance and reliability of LLMs in generating accurate information are significantly influenced by multiple factors, including training data quality, model architecture design, and post-training optimization processes \cite{naveed2024comprehensiveoverviewlargelanguage}, \cite{minaee2024largelanguagemodelssurvey}, \cite{guo2023evaluatinglargelanguagemodels}.

However, a significant challenge in the deployment of LLMs lies in their propensity to generate nonfactual responses, a phenomenon commonly referred to as hallucination. These hallucinations fundamentally undermine the reliability and faithfulness of LLMs, presenting substantial obstacles to their widespread adoption across various domains \cite{Huang_2024}, \cite{sahoo2024comprehensivesurveyhallucinationlarge}. The mitigation of hallucinations has consequently emerged as a critical area of research within the field. While various strategies have been proposed and implemented to reduce hallucinations, showing promising improvements in the faithfulness of LLMs for general-purpose tasks, domain-specific applications remain particularly challenging \cite{tonmoy2024comprehensivesurveyhallucinationmitigation}, \cite{rawte-etal-2023-troubling}, \cite{berberette2024redefininghallucinationllmspsychologyinformed}.

In this paper, we present a comprehensive study evaluating the extent of hallucination in LLMs under domain-specific prompting, with a particular focus on scientific literature. We develop and implement a systematic evaluation pipeline to assess fifteen prominent open-source LLMs: Qwen 2.5 \cite{qwen2}, Gemma 2 \cite{gemmateam2024gemma2improvingopen}, Llama 3 \cite{grattafiori2024llama3herdmodels}, Phi 3 \cite{abdin2024phi3technicalreporthighly}, Orca 2 \cite{mitra2023orca2teachingsmall}, Mistral v-0.3 \cite{mistral}, Deepseek-llm \cite{deepseekai2024deepseekllmscalingopensource}, Olmo-2 \cite{olmo20242olmo2furious}, Mistral-Nemo \cite{mistrala56:online}, Eurus-2 \cite{yuan2024implicitprm}, and Solar-Pro \cite{upstages88:online}. Our evaluation utilizes the ArXiv dataset \cite{clement2019usearxivdataset} as the primary source of scientific articles, providing a robust foundation for assessing model performance in academic contexts.

The evaluation pipeline \textbf{ArxEval }introduces two novel tasks: \textbf{Jumbled Titles} and \textbf{Mixed Titles}. These tasks are specifically designed to assess the faithfulness of LLMs in retrieving and reasoning about scientific articles under challenging conditions. The models are presented with either jumbled or mixed titles and evaluated not only on their prompt adherence but also on the quality and accuracy of their generated outputs. By adopting an open-ended evaluation approach, we aim to provide comprehensive insights into the models' capabilities in processing and responding to ambiguous or altered inputs within a domain-specific context, particularly focusing on their ability to maintain factual accuracy while handling complex scientific information.

This study contributes to the growing body of research on LLM reliability and provides valuable insights into the current limitations and capabilities of state-of-the-art language models in handling domain-specific tasks. Our findings have important implications for the development and deployment of LLMs in scientific and academic applications, where maintaining factual accuracy is crucial.

\section{Related Work}

\subsection{Hallucinations in Large Language Models (LLMs)}
Hallucinations in LLMs have been extensively studied and documented. While significant advancements have been made in improving their accuracy and reliability, LLMs have been shown to hallucinate even when tasked with generating responses based on known facts \cite{jiang2024large}. Such behavior suggests an inherent limitation in these models, reinforcing the hypothesis that hallucination may be an intrinsic characteristic \cite{banerjee2024llms}.

\subsection{Hallucinations in Domain-Specific Settings}

\subsubsection{Definition and Challenges}
Domain-specific hallucinations manifest when LLMs generate inaccurate or fabricated information in specialized fields. In domains like biomedicine, such hallucinations can have serious implications, potentially leading to incorrect medical advice or misinterpretation of research data. The fundamental challenge lies in maintaining factual accuracy while preserving the model's ability to generate coherent and contextually relevant responses.

\subsubsection{Causes and Perspectives}

Domain-specific hallucinations primarily stem from two factors: deficiencies in training data and limitations in model architecture \cite{dziri-etal-2022-origin}. However, recent research presents an alternative viewpoint, suggesting that under certain conditions, hallucinations could be leveraged as a resource for novel problem-solving approaches \cite{sui2024confabulation}.

\subsubsection{Detection and Evaluation Frameworks}
\begin{itemize}
    \item \textbf{DelucionQA} \cite{sadat2023delucionqadetectinghallucinationsdomainspecific}: A specialized dataset designed for detecting hallucinations in domain-specific question-answering tasks, providing evaluation metrics for retrieval-augmented LLMs.
    \item \textbf{DAHL} \cite{seo2024dahldomainspecificautomatedhallucination}: A comprehensive benchmark for evaluating hallucinations in biomedical text generation, featuring atomic unit decomposition and the DAHL Score metric.
\end{itemize}

\subsection{Hallucinations in Multimodal Settings}

\subsubsection{Definition and Challenges}
Multimodal hallucinations occur when models generate outputs inconsistent with visual or auditory inputs. This phenomenon is particularly critical in applications like video understanding, where temporal and spatial accuracy are essential \cite{bai2024hallucinationmultimodallargelanguage}.

\subsubsection{Evaluation Frameworks}
\begin{itemize}
    \item \textbf{VidHalluc} \cite{li2024vidhallucevaluatingtemporalhallucinations}: A specialized benchmark for evaluating temporal hallucinations in video understanding, assessing multiple dimensions including action recognition and scene transitions.
    \item \textbf{MHaluBench} \cite{chen-etal-2024-unified-hallucination}: A meta-evaluation framework for comprehensive multimodal hallucination detection across diverse categories.
\end{itemize}

\subsection{Hallucinations in Natural Language Generation}
In natural language generation tasks such as dialogue generation, abstractive summarization, and neural machine translation, hallucinations frequently manifest as plausible but factually incorrect outputs \cite{Ji_2023}. These inaccuracies significantly impact the reliability and trustworthiness of these models.
\subsection{Hallucinations in Academic Reference Generation}
Academic reference generation represents a critical challenge, with studies demonstrating that even state-of-the-art models frequently generate fabricated or inaccurate citations \cite{agrawal-etal-2024-language}. This limitation underscores the urgent need for continued research in hallucination mitigation, particularly in tasks where factual accuracy is paramount.
In addressing these challenges, our work specifically focuses on \textit{Domain-Specific Biases} by leveraging over 150 categories of papers from the ArXiv repository, providing a comprehensive evaluation across diverse academic domains.

\begin{figure}[h!]
    \centering
    \includegraphics[width=0.7\linewidth]{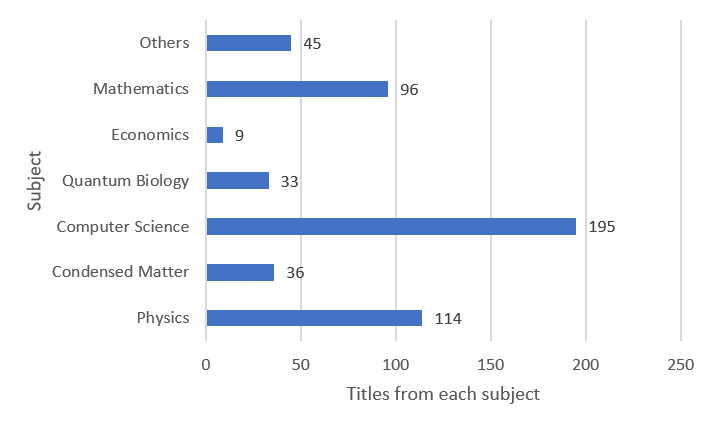}
    \caption{Number of titles from each subject.}
    \label{f1}
\end{figure}

\section{Dataset}

Here, we describe the datasets used for the \textbf{Jumbled Title} task and the \textbf{Mixed Title} task. A total of 176 categories were used to construct the dataset for both tasks. We used 3 titles from each category giving us 528 titles. Figure \ref{f1} shows the number of titles from each subject, Computer Science had the most number of categories at 65 having a total of 195 titles. Economics had the least at 3 categories and 9 titles. 

In Table \ref{t3} we provide a further insights into the dataset for both the Jumbled Titles and Mixed Titles tasks. Using the Flesch reading ease score \cite{Flesch–K46:online}, we find that Jumbled Titles falls in the \texttt{Very difficult to read. Best understood by university graduates} range, while Mixed Titles is categorized as \texttt{Extremely difficult to read. Best understood by university graduates}. Using the Gunning fog index \cite{Gunningf10:online}, we find Jumbled Titles scoring 17 falls into the \texttt{College graduate} level. Mixed Titles scores 19, which exceeds the standard Gunning fog categories but is most appropriately classified as \texttt{College graduate} level. These scores further validate the technical complexity of our dataset. The dataset exhibits diverse title lengths, with Jumbled Titles ranging from 2-24 words and Mixed Titles from 8-33 words. This broad distribution ensures robust evaluation across varying input complexities.

The \textbf{Jumbled Title} task includes 528 jumbled titles, distributed among the 176 categories of the ArXiv dataset, Table \ref{t1} presents a few examples of the dataset used for the Jumbled Title task.

\begin{table}[h!] 
    \centering 
    \begin{tabular}{|p{0.45\linewidth}|p{0.45\linewidth}|}
        \hline
        \textbf{Jumbled Title} & \textbf{Original Title} \\
        \hline
        Hydrodynamic bubble to obstruction expansion & Hydrodynamic obstruction to bubble expansion \\ \hline
        with Warm Microwave Background Constraining Inflation Cosmic the & Constraining Warm Inflation with the Cosmic Microwave Background \\ \hline
        QCD Hadron Colliders Three-Jet Corrections Production Two-Loop at for Leading-Color & Leading-Color Two-Loop QCD Corrections for Three-Jet Production at Hadron Colliders \\ \hline
        enumeration theorem polynomials Order P\'olya's and & Order polynomials and P\'olya's enumeration theorem \\
        \hline
    \end{tabular}
    \caption{Examples from the dataset used for the Jumbled Title task}
    \label{t1}
\end{table}

The \textbf{Mixed Title} task consists of 265 mixed titles, derived from the 176 categories in the ArXiv dataset. Table \ref{t2} shows examples from the dataset used for the Mixed Title task.

\begin{table*}[h!]
    \centering
    \begin{tabular}{|p{6cm}|p{5cm}|p{5cm}|}
        \hline
        \textbf{Mixed Title} & \textbf{Title 1} & \textbf{Title 2} \\ \hline
        Bioconvection Transport Irradiation Suspensions: across Non-scattering Coarse-grain Molecular under Heating Fullerene in Membrane Collimated Above a Study Phototactic from Cell Dynamics of & Heating from Above in Non-scattering Suspensions: Phototactic Bioconvection under Collimated Irradiation & Coarse-grain Molecular Dynamics Study of Fullerene Transport across a Cell Membrane \\ \hline
        Value of oil and gas Semi-intrusive exchange in the uncertainty quantity multiscale for stock gas london and change disclosures components oil of the relevance models propagation of upstream reserve companies & Value relevance of the components of oil and gas reserve quantity change disclosures of upstream oil and gas companies in the London Stock Exchange & Semi-intrusive uncertainty propagation for multiscale models \\ \hline
    \end{tabular}
    \caption{Overview of the dataset for the Mixed Title task}
    \label{t2}
\end{table*}

\begin{table}[h!]
    \centering
    \begin{tabular}{|l|c|c|c|c|c|c|}
    \hline
    \textbf{Task} & \textbf{Total} & \textbf{Categories Count} & \textbf{Avg. Len} & \textbf{Range} & \multicolumn{2}{c|}{\textbf{Readability}} \\
    \cline{6-7}
    & & & & & \textbf{Gunning Fog} & \textbf{Flesch} \\
    \hline
    Jumbled Titles & 528 & 176(3) & 9.45 & 2-24 & 17 & 16 \\
    \hline
    Mixed Titles & 265 & 176(3) & 18.89 & 8-33 & 20 & 8 \\
    \hline
    \end{tabular}
    \caption{Dataset Statistics for Jumbled and Mixed Titles Tasks}
    \label{t3}
\end{table}

\section{Methodology}

\begin{figure*}[!ht]
    \centering
    \includegraphics[width=0.9\columnwidth]{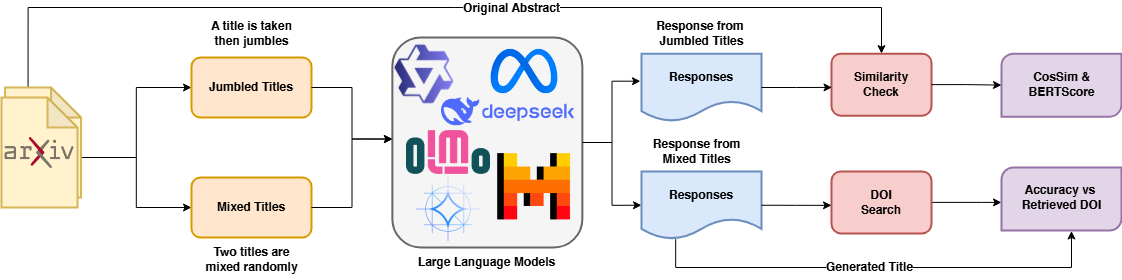}
    \caption{Our proposed pipeline for evaluating the LMs on use of the ArXiv as source for papers.}
    \label{f2}
\end{figure*}

Our pipeline consists of two main tasks as seen in Figure \ref{f2}. With our prompts, we aim to demonstrate and experiment with how users would typically interact with these models in real-world scenarios. This approach reflects the reality that average users, and even expert users, rarely invest time in fine-tuning their prompts with elaborate instructions during routine interactions.

\subsection{Jumbled Titles}

We select 5 titles from each category available in the ArXiv dataset. Each title is scrambled within itself, creating a jumbled version of the original title. These jumbled titles serve as input to the LLMs, which are prompted using the following template:\\
\begin{mdframed}
\texttt{Tell me about this research paper:} \textit{[title]}\\
\end{mdframed}
Where \textit{[title]} represents the jumbled title.

Algorithm \ref{a1} describes how the dataset is built for the Jumbled Titles task. Once the language models generate their responses, we evaluate the similarity of these responses against the original abstract of the corresponding paper. This evaluation is performed using the \textit{all-MiniLM-L6-v2} model [\cite{reimers2019sentencebertsentenceembeddingsusing}] to generate embeddings. To quantify the degree of alignment between the generated response and the original abstract, cosine similarity scores are calculated. We also use BERTScore[\cite{zhang2020bertscoreevaluatingtextgeneration}] and Semantic Textual Similarity (STS) \cite{reimers2019sentencebertsentenceembeddingsusing} as additional metrics for evaluating similarities. 

\begin{algorithm}
\caption{Create Jumbled Titles Dataset}
\begin{algorithmic}[1]
\Require Parquet file path \textit{parquet\_file\_path}, Output CSV file path \textit{output\_csv\_file\_path}
\Ensure CSV file with jumbled titles is saved

\Function{jumble\_title}{title}
    \State Split the \textit{title} into words
    \State Randomly shuffle the words
    \State Return the shuffled words joined into a single string
\EndFunction

\Procedure{create\_jumbled\_titles\_dataset}{parquet\_file\_path, output\_csv\_file\_path}
    \State Load the dataset from \textit{parquet\_file\_path} into DataFrame \textit{df}
    \State Apply the \textit{jumble\_title} function to the \textit{title} column in \textit{df}
    \State Create a new DataFrame \textit{jumbled\_titles\_df} with the jumbled titles
    \State Save \textit{jumbled\_titles\_df} to CSV file at \textit{output\_csv\_file\_path} without the index
\EndProcedure

\end{algorithmic}
\label{a1}
\end{algorithm}

\subsection{Mixed Titles}

In the second part of the pipeline, we create mixed titles by combining two random paper titles from the dataset. These mixed titles are then used as input to the LLMs with the following prompt template:\\

\begin{mdframed}
    \texttt{Tell me 2 papers related to this and only mention the Title and the DOI:} \textit{[title]}\\

\end{mdframed}

Where \textit{[title]} represents the mixed title.

Algorithm \ref{a2} shows how the Mixed Titles dataset is built. After the language models generate their responses, we perform a detailed evaluation of the provided DOIs using the \textit{Crossref API}, \textit{DataCite API}, \textit{UnPaywall API} and \textit{OpenAlex API} to ensure thorough search of all possible DOIs is done. The evaluation process involves the following two steps:

\begin{enumerate}
    \item \textbf{Checking DOI Validity:} For each model-generated DOI, we verify its existence through API requests to established academic databases. This validation step is crucial for assessing the model's ability to generate legitimate academic identifiers.
    \item \textbf{Verifying Title Accuracy:} For validated DOIs, we retrieve the official paper titles from the corresponding databases and compare them against the model-generated titles. This comparison enables us to evaluate the model's accuracy in associating DOIs with their correct academic works.
\end{enumerate}

\begin{algorithm} [!ht]
\caption{Create Mixed Titles Dataset}
\begin{algorithmic}[1]
\Require Parquet file path \textit{parquet\_file\_path}, Output CSV file path \textit{output\_csv\_file\_path}
\Ensure CSV file with mixed titles is saved

\Function{mix\_titles}{title1, title2}
    \State Split \textit{title1} and \textit{title2} into words
    \State Concatenate the words from both titles into a list \textit{mixed\_words}
    \State Randomly shuffle \textit{mixed\_words}
    \State Return the shuffled words joined into a single string
\EndFunction

\Procedure{create\_mixed\_titles\_dataset}{parquet\_file\_path, output\_csv\_file\_path}
    \State Load the dataset from \textit{parquet\_file\_path} into DataFrame \textit{df}
    \State Extract the \textit{title} column from \textit{df} and convert it to a list \textit{titles}
    \State Randomly shuffle the \textit{titles} list
    \If {the length of \textit{titles} is odd}
        \State Append an empty string to \textit{titles}
    \EndIf
    \State Initialize an empty list \textit{mixed\_titles\_data}
    \For {each pair of titles in \textit{titles}}
        \State Mix the pair of titles using the \textit{mix\_titles} function
        \State Append a dictionary with keys \textit{mixed\_title}, \textit{title1}, and \textit{title2} to \textit{mixed\_titles\_data}
    \EndFor
    \State Create a new DataFrame \textit{mixed\_titles\_df} from \textit{mixed\_titles\_data}
\EndProcedure

\end{algorithmic}
\label{a2}
\end{algorithm}

\begin{figure}[!ht]
    \centering
    \includegraphics[width=0.7\linewidth]{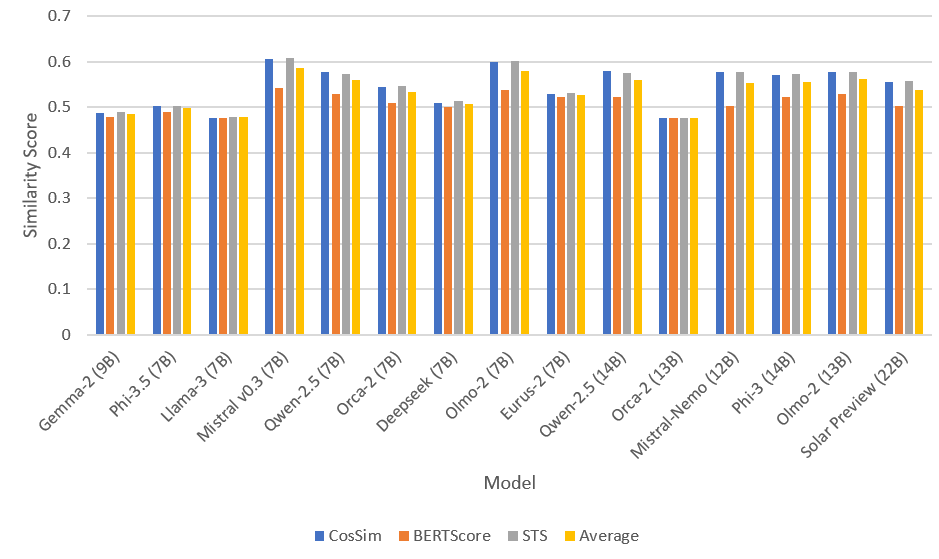}
    \caption{CosSim, BERTScore and STS Scores for all models.}
    \label{f3}
\end{figure}

\begin{table}[h!]
    \centering
    \begin{tabular}{|l|c|c|c|c|c|}
    \hline
    \textbf{Model} & \textbf{Parameters} & \textbf{CosSim} & \textbf{BERTScore} & \textbf{STS} & \textbf{Average} \\
        \hline
        Gemma-2        & 9B       & 0.487     &   0.478  & 0.489 & 0.485 \\
        Phi-3.5        & 7B       & 0.502     &   0.489  & 0.503 & 0.498 \\
        Llama-3        & 7B       & 0.477     &   0.477  & 0.479 & 0.478 \\
        Mistral v0.3   & 7B       & \textbf{0.605}&   \textbf{0.542} & \textbf{0.607} & \textbf{0.585} \\
        Qwen-2.5       & 7B       & 0.578     &   0.528  & 0.572 & 0.559 \\
        Orca-2         & 7B       & 0.544     &   0.508  & 0.546 & 0.533 \\
        Deepseek       & 7B       & 0.509     &   0.501  & 0.513 & 0.507 \\
        Olmo-2         & 7B       & \textit{0.600} &\textit{0.537} & \textit{0.602} & \textit{0.580} \\
        Eurus2         & 7B       & 0.528    &   0.523   & 0.530 & 0.527 \\
        Qwen-2.5       & 14B      & 0.579     &   0.522  & 0.575 & 0.559 \\
        Orca-2         & 13B      & 0.476     &   0.475  & 0.477 & 0.476 \\
        Mistral-Nemo   & 12B      & 0.577     &   0.503  & 0.578 & 0.553 \\
        Phi-3          & 14B      & 0.571     &   0.522  & 0.572 & 0.555 \\
        Olmo-2         & 13B      & 0.577     &   0.528  & 0.578 & 0.561 \\
        Solar Preview  & 22B      & 0.55      &   0.502  & 0.558 & 0.537 \\
        \hline
    \end{tabular}
    \caption{Cosine Similarity Scores, BERTScores and STS Scores between generated response and the original abstract of the paper for various language models. Best performing model represented by \textbf{Bold} and second best by \textit{Italics}.}
    \label{t4}
\end{table}

\begin{figure}
    \centering
    \includegraphics[width=0.7\linewidth]{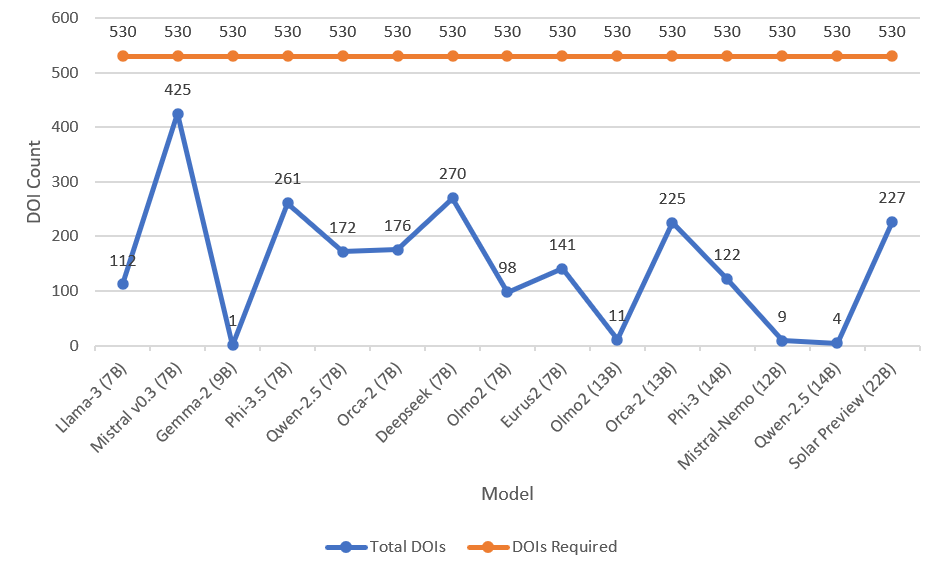}
    \caption{DOIs generated by each model when prompted during the \textit{Mixed Title} task.}
    \label{f4}
\end{figure}

\section{Results}

To run the inference on the models we used $2\times$ T4 Tesla [16GB]. We made use of PyTorch \cite{paszke2019pytorchimperativestylehighperformance} and Huggingface's Transformers \cite{wolf2020huggingfacestransformersstateoftheartnatural}. It took us approximately 2.5 to 3 hrs on average to run the entire pipeline for each model. We also use 4-bit quantization using bitsandbytes \cite{bitsandb} for faster and efficient inference.

In Table \ref{t4}, we present the performance of models on the Jumbled Titles task. Mistral v0.3 \cite{jiang2023mistral7b} achieved the highest similarity scores, with 0.605 on cosine similarity, 0.542 on BERTScore and 0.607 on STS. This was followed by Qwen2.5 (7B) being the second best performing model overall. On average, BERTScore reduced the similarity between the texts by 1.068\% compared to cosine similarity. The worst performing model on all tasks was Orca-2 (13B) with cosine similarity, BERTScore and STS at 0.476, 0.475, 0.477 respectively averaging 0.476.

\begin{table*}[!ht]
\centering
    \begin{tabular}{|l|c|c|c|c|}
        \hline
        \textbf{Model} & \textbf{Total DOIs} & \textbf{DOIs Found} & \textbf{DOIs Not Found} & \textbf{Matching Titles} \\
        \hline
        Llama-3      (7B)& 112  & 9 [8.04\%]  & 8 [91.96\%]  & 0.00\% \\
        Mistral v0.3 (7B)&  425 & 109 [25.65\%]  & 316 [74.35\%]  & 0.00\% \\
        Gemma-2      (9B)& 1   & 0 [0.00\%]      & 1 [100.00\%]    & 0.00\%     \\
        Phi-3.5      (7B)& 261 & 36 [13.79\%] & 225 [86.21\%]& 0.00\% \\
        Qwen-2.5     (7B)& 172  & 70 [40.70\%]    & 102 [59.30\%] & 0.00\% \\
        Orca-2       (7B)& 176 & 20 [18.87\%] & 86 [81.13\%] & 0.00\% \\
        Deepseek     (7B)& 270 & 62 [22.96\%] & 208 [77.04\%]& 0.00\% \\
        Olmo2        (7B)& 98 & 12 [12.24\%]    & 86 [87.76\%]& 0.00\% \\
        Eurus2        (7B)& 141 & 40 [28.37\%]    & 101 [71.63\%]& 0.00\% \\
        Olmo2        (13B)& 11 & 4 [36.36\%]    & 7 [63.64\%]& 0.00\% \\
        Orca-2       (13B)& 225 & 39 [17.33\%] & 186 [82.67\%] & 0.00\% \\
        Phi-3        (14B)& 122 & 35 [28.69\%] & 87 [71.31\%] & 0.00\% \\
        Mistral-Nemo (12B)& 9  & 2 [22.22\%]  &  7 [77.78\%] & 0.00\% \\
        Qwen-2.5     (14B)&  4  & 0 [0.00\%]  & 4 [100.00\%] & 0.00\% \\
        Solar Preview(22B)& 227 & 59 [25.99\%]& 168 [74.01\%] & 0.00\% \\
        \hline
    \end{tabular}
    \caption{DOI Search and Title Comparison Results}
    \label{t5}
\end{table*}

\begin{figure}[!ht]
    \centering
    \includegraphics[width=0.7\linewidth]{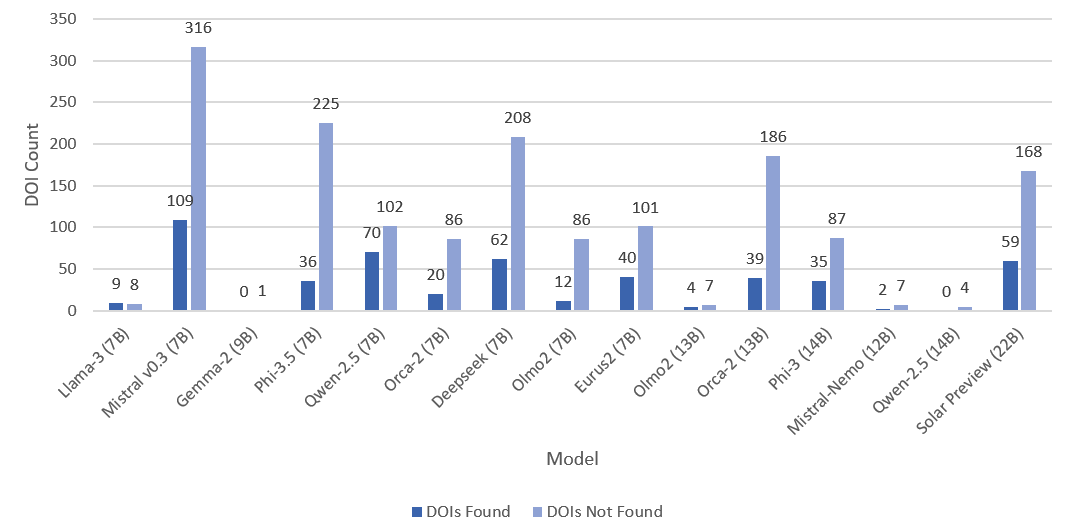}
    \caption{DOIs Found for each model vs DOIs not for each model}
    \label{f5}
\end{figure}

In Table \ref{t5}, we evaluate the performance of the models on the Mixed Titles task. Among the models, Mistral-v0.3 (7B) generated the highest number of DOIs at 425, with 25.56\% of the generated DOIs actually existing. Gemma-2 (9B) and Qwen-2.5 (14B) had the lowest percentage of valid DOIs, with 0\% of the generated DOIs found to exist. In contrast, the model with the highest percentage of valid DOIs was the smaller Qwen-2.5 (7B), achieving a 40.70\% validity rate with 70 valid DOIs generated. In Figure \ref{f5}, we see a representation of the DOIs found vs DOIs not found for each model; in all cases, the number of DOIs not found is higher than that of the DOIs found. All models demonstrated a significant shortcoming: for every valid DOI retrieved, the associated title was incorrect 100\% of the time when compared to the title generated by the model. It is also worth mentioning that Mistral-v0.3(7B) generated the most number of valid DOIs at 109 of the 425 generated actually existing

\section{Conclusion}\label{con}

This paper evaluates the extent of hallucination in state-of-the-art language models by designing two tasks: \textbf{Jumbled Titles} and \textbf{Mixed Titles}. In the Jumbled Titles task, the fifteen evaluated models achieved an average cosine similarity score of \textsc{0.544}, \textsc{0.509} on BERTScore, and \textsc{0.545} on STS. Mistral-v0.3 was the best-performing model across all metrics averaging \textsc{0.585} on the Jumbled Titles task, outperforming models twice and thrice its size.

For the Mixed Titles task, while models generated DOIs for the mixed titles, they often cited non-existent papers or mismatched DOIs. These results underscore critical limitations in maintaining factual accuracy in domain-specific contexts. On average, valid DOIs were generated only \textsc{17.75\%} of the time. Every model also completely failed to generate the corresponding DOI for the title they generated, showing that models struggle with maintaining factual consistency.
To further highlight \textit{Prompt Adherence}, it is worth noting that none of the models generated the required number of two DOIs for each \textit{Mixed Title}. This results in a discrepancy, as the expected output for each model was 530 DOIs (given 265 mixed titles), but none of the models met this requirement, as seen in Figure \ref{f4}.

\subsection{Model Size Performance}\label{ems}
In Table \ref{t4} and Table \ref{t5}, we observe a concerning trend where many of the larger models are significantly outperformed by their smaller counterparts in both tasks. For instance, the 7B Mistral-v0.3 outperforms models up to three times its size, while the Solar Preview(22B) demonstrates mediocre performance despite its substantially larger parameter count, even for the same models with Qwen2.5 where the 7B parameter model outperforms the 14B parameter variant. This pattern raises serious questions about the relationship between model size and task performance. The results starkly highlight that simply scaling up model parameters does not guarantee better performance in specialized tasks, particularly those requiring precise adherence to instructions and factual accuracy. This counterintuitive finding challenges the common assumption that larger language models inherently perform better, suggesting that architectural choices and training approaches might be more crucial than raw parameter count for achieving superior performance.

\section{Limitations}

There are several limitations to our work:

\begin{enumerate}
    
    \item \textbf{Model Selection}: Our evaluation focuses on smaller models due to computational constraints. Results may differ significantly with larger variants, which could exhibit different performance characteristics. Although our findings in Section \ref{ems} shows that might not always be the case, as we observed smaller models often outperforming their larger counterparts.
    
    \item \textbf{Model Quantization}: We use 4-bit quantization for inference. While this may reduce performance, studies suggest the impact is minimal \cite{jin2024comprehensiveevaluationquantizationstrategies}. The trade-off between computational efficiency and potential performance impact was deemed acceptable for our experimental setup.
    
\end{enumerate}

\bibliographystyle{unsrt}  
\bibliography{references}

\end{document}